\documentclass[letterpaper]{llncs}
\usepackage[letterpaper]{geometry}
\usepackage{makeidx}  

\usepackage[T1]{fontenc} 
\usepackage[utf8]{inputenc}
\usepackage{amsmath, amssymb, graphicx, url, algorithm2e}
\DeclareUnicodeCharacter{00A0}{ } 

\renewcommand{\vec}[1]{\boldsymbol{\mathbf{#1}}}
\newcommand{\mat}[1]{\vec{#1}}

\newcommand{\E}[1]{\left\langle #1 \right\rangle}

\newcommand{\N}{\mathcal{N}}

\newcommand{\Multinomial}{\mathrm{Multinomial}}

\newcommand{\Bernoulli}{\mathrm{Bernoulli}}
\newcommand{\Beta}{\mathrm{Beta}}

\newcommand{\W}{\mat{W}} 
\newcommand{\w}{\vec{w}} 
\newcommand{\z}{\vec{z}} 
\newcommand{\Z}{\mat{Z}} 
\newcommand{\x}{\vec{x}} 
\newcommand{\X}{\mat{X}} 
\newcommand{\I}{\mat{I}} 

\newcommand{\R}{\ensuremath{\mathbb R}}

\usepackage{subfigure}

\begin{document}

\mainmatter              
\title{Classification of weak multi-view signals by sharing factors in a mixture of Bayesian group factor analyzers}
\titlerunning{Sharing factors in a mixture of GFAs}  
%
\author{Sami Remes\inst{1} \and Tommi Mononen\inst{1,2}
\and Samuel Kaski\inst{1}}
\authorrunning{Sami Remes, Tommi Mononen and Samuel Kaski} 

\institute{Helsinki Institute for Information Technology HIIT, Department of Computer Science, Aalto University \and Department of Neuroscience and Biomedical Engineering, Aalto University}

\maketitle              

\begin{abstract}
We propose a novel classification model for weak signal data, building upon a recent model for Bayesian multi-view learning, Group Factor Analysis (GFA).
Instead of assuming all data to come from a single GFA model, we allow latent clusters, each having a different GFA model and producing a different class distribution. We show that sharing information across the clusters, by sharing factors, increases the classification accuracy considerably; the shared factors essentially form a flexible noise model that explains away the part of data not related to classification. Motivation for the setting comes from single-trial functional brain imaging data, having a very low signal-to-noise ratio and a natural multi-view setting, with the different sensors, measurement modalities (EEG, MEG, fMRI) and possible auxiliary information as views. We demonstrate our model on a MEG dataset.

\end{abstract}

\section{Introduction}

Recently a lot of focus in machine learning has been given to the analysis of multi-view data, a scenario in which the data come from multiple data sources. One observation or data point consists of data from multiple sources, and in this sense each data source can be considered a different ``view'' to the same ``object''. This is a natural setting with, e.g. brain imaging data where each sensor, measurement modality (including EEG, MEG and fMRI) and possible auxiliary or experimental information form the views. Notably, the relevant, discriminating signal between, e.g., different experimental conditions is very weak compared to all other activity on-going in the brain.

There has been increasing interest to explore neuroscientific data using machine learning methods, which are more capable to reveal the complex phenomena happening in the brain than traditional contrasting methods (e.g. t-tests). A popular application area of machine learning is called brain decoding. The goal is to infer, based on a given brain signal, what task a subject is performing, given a training set consisting of sample signals of performed tasks. These samples are usually from single trials that have a low signal-to-noise ratio. 

However, good prediction performance does not necessarily imply increase of neuroscientific knowledge about the processes in the brain. Some of the best prediction methods, such as SVM and Gaussian processes, are essentially black boxes, and it is hard to infer what the predictions are based on. Neuroscientists are interested in learning what parts of the brain have an effect on the prediction. 
Therefore, prediction methods that do not give these plausible explanations are much more difficult to use for accumulating neuroscientific knowledge. 
Generative probabilistic models are generally more immediately interpretable \cite{kiamulti,haufe2014}, and for instance when using linear generative models, the weights point to those brain locations that are activated due to a performed task.

A MEG device (as well as EEG) outputs a very fine-scale time series. Some methods such as the linear discriminant analysis and its various regularized versions that are widely used in brain-computer interfaces require that this rich source of information is squeezed into a single value (e.g. maximum or mean) \cite{lemm2011introduction,hohne2014mean}. 
Methods should take advantage of the time structure, otherwise useful information might get lost altogether or attributed to wrong time points.

We propose a novel Bayesian multi-view generative classification model that takes the time information into account in an intuitive way, by directly modelling the time dependencies. The generative approach is generally beneficial in scenarios where training data is scarce, because the assumptions of the generative process help learning the data efficiently. The model is motivated by setups common in brain imaging data, but applicability of the proposed model is not restricted only to neuroscience. The model can be applied to any time-varying or otherwise structured multiple-data-source framework. 

Recently, unsupervised multiple-data-source modelling in the Bayesian framework has been studied, a state-of-the-art solution being the group factor analysis (GFA) model \cite{Virtanen12aistats,gfajournal}. Similar models have been studied also with different priors for sparsity \cite{zhao2014bayesian} and using an optimization-based sparse dictionary coding approach \cite{jia2010factorized}. 
We propose learning a number of GFA models per class label, a mixture modelling approach commonly used for classification (e.g., in voice recognition with GMMs \cite{reynolds1995speaker}). We jointly learn a set of clusters and their respective label distributions rather than learning models separately for each label.
Additionally, we assume that the GFA models share some of their factors, which turns out very significant for classification. The large common parts of the signals can be modelled and explained away with the shared factors, while the weak and likely the most interesting discriminating parts of the signal can be modelled by the cluster-specific parameters.

Other approaches to multi-view learning also exist. A CCA-type approach was used where samples in multiple views are transformed to a common discriminative space where within-class variation is minimized while between-class variation is maximized for all views \cite{kan2012multi}. Approaches in \cite{ek2008gaussian} and \cite{eleftheriadis2013shared} find a common latent space for the views using Gaussian processes. Optimization-based approaches \cite{wang2013multi,zhang2012joint}  regress the data to the clusters and class labels, but they are not doing generative modelling in a way related to our GFA based approach. In \cite{santana2015multi} the authors studied using different
types of features and applied classifiers with all the different subsets of the feature types, and
showed that using a combination of features improved classification accuracy compared to a single feature type.

In the experimental section, we apply our model on MEG data and compare it against a group LASSO classifier \cite{yuan2006model} that is known to generally perform well, to the extent of being hard to beat in practice. The classifier takes the multi-source nature of the data into account in the same way as our model does, making the Group LASSO a natural baseline.

\section{Model}

First we present a GFA model for multi-view data, containing multiple data sources of possibly different dimensionalities. 
Next we extend GFA to a mixture model where some factors are shared between the mixture components (clusters). 
This effectively separates the parts of data that contribute to differences between the classes from the rest, allowing to share statistical strength to learn the shared parts more efficiently. Finally, the model is extended to include class labels. 

\subsection{Group Factor Analysis}

Group factor analysis \cite{gfajournal,Virtanen12aistats} can be seen both as an extension of Bayesian CCA
\cite{Klami13jmlr} to multiple data sources, or alternatively it can be
seen as an extension of factor analysis that treats the data sources
similarly to how regular factor analysis treats individual variables.

Let $\x_n^{(m)} \in \R^{D_m}$ denote the $n$th sample (out of the data size $N$) of the $m$th
data source, the model likelihood is given by
\begin{equation}
    \x_n^{(m)} | \W,\z_n,\tau_m \sim \N( \W^{(m)}\z_n, \tau_m^{-1}\I ) \; .
    \label{eq:gfa}
\end{equation}
The observations (data points) are modeled with unknown latent
variables $\z_n \in \R^K$ corresponding to $K$ factors, which are then
mapped to the data space with data source specific linear projections $\W^{(m)}$.
The latent variables are shared between all
data sources and can therefore model correlations between them, via 
controlling the values of $\W^{(m)}$ with a specific type of sparsity
constraint; if some factor $k$ is not useful for modelling the source $m$, we
want $\w^{(m)}_k$ (the $k$th column of $\W^{(m)}$) to be zero. The
desired structure is achieved with the automatic
relevance determination (ARD) prior
\begin{equation}
    \w_k^{(m)} | \alpha_k^{(m)} \sim \N( \vec0, (\alpha_k^{(m)})^{-1}\I ),
    \label{eq:gfa-W}
\end{equation}
where the $\alpha$-parameters have independently non-informative gamma priors
$\alpha_k^{(m)} \sim \Gamma(10^{-14},10^{-14})$. 

\subsection{Mixture of GFAs}

We next describe a mixture of GFAs model that will be further extended later in this section. In our application scenario we assume that a single GFA model could explain most of the variation in the data, but not the most interesting part which is assumed to be weak. We assume the weak signal to consist of distinct parts (clusters), each of which can also be modeled by a set of factors; the clusters can consist of experimental conditions, for instance. 
The data is therefore generated by both cluster-specific factors $\W_c$, as well as shared factors $\hat\W$ (active regardless of cluster assignment) as follows:
\begin{equation}
    \x_n^{(m)}|c_n \sim \N(\W^{(m)}_{c_n}\z_n + \hat\W^{(m)}\hat\z_n, (\tau_{c_n}^{(m)}\I)^{-1}),
    \label{}
\end{equation}
where a latent variable $c_n$ indexes the cluster which the data point belongs to. The first term in the normal mean specifies the cluster-specific parts of data and the latter term the parts shared by the clusters (information not discriminating classes). Factor loadings have group-sparse ARD priors
\begin{align}
    [\W_{c}^{(m)}]_k &\sim \N(\vec0,(\alpha_{c,k}^{(m)})^{-1}\I),
\end{align}
where the factors are indexed with $k\in\{1,\ldots,K\}$ and clusters with $c\in\{1,\ldots,S\}$.
The shared loadings have similarly
$\hat\W^{(m)}_k\sim\N\left(\vec0,\hat\alpha_k^{-1}\I\right)$
but here we used different hyperparameters for the ARD. Since the signal is known to be weak, a model that explains all data by shared components would be almost equally good, and therefore we need to tell the model to prefer solutions having some cluster-specific components. That can be achieved by setting the hyperparameter $a$ for the shared components to a larger value, such as $a=30$ in our experiments. This brings the prior mean to  $\E{\alpha}  = a/b \gg 1$ while still having a large variance. We recommend setting the value large enough such that during inference both shared and cluster-specific factors are found.

The cluster assignments $c_n$ for each sample $n$ are given by $c_n | \vec\pi \sim \Multinomial(\vec\pi)$
with a conjugate Dirichlet prior for the prior probabilities $\vec\pi$.
Noise precisions $\tau_c^{(m)}$ have non-informative Gamma priors.
The values of $\tau$ determine how much of the uninteresting signal can be explained away by the ``simple'' noise model and how much needs to be explained by the shared components. 

\subsection{Classifying GFA mixture}

We extend the GFA mixture model to include a part that ties the
mixture components also to the class labels. Here we consider only two classes, although it is straightforward to extend our model into a multi-class setting. Let  $r_n$ be our observed class label (0 or 1) and  $c_n$ be latent cluster assignments. We model the output label $r_n$ as a binary variable with a Bernoulli distribution
$r_n | c_n,\vec\gamma \sim \Bernoulli(\gamma_{c_n})$, 
where we use the conjugate prior $\gamma_{c} \sim \Beta(\frac{1}{2},\frac{1}{2})$ that denotes the probability $P(r_n = 1 | c_n = c)$.

The full joint density for the model is given by
\begin{align}
p(\Theta|\X, \vec r) \propto p(\X|\W,\Z,\vec c,\vec\tau)p(\vec r|\vec\gamma,\vec c)^\beta p(\vec c|\vec\pi)p(\vec\pi)p(\vec\gamma)p(\Z)p(\vec\tau)p(\W|\vec\alpha)p(\vec\alpha),
\end{align}
where we have given an additional weight $\beta$ to the Bernoulli likelihood of the output labels
to facilitate better classification results by making the clusters more likely to match the labels. In our experiments we set $\beta = 100$. We recommend to tune the parameter large enough such that the clustering given by the model, starts to match well with the class labels in a training set. Here we also used short-hand notations
$\X$ for the data matrix, $\W$ for all loading matrices and $\Z$ for all latent variables, $\vec c$ for the cluster assignments $c_n$, and $\vec r$ for the output labels.

Given the model specification as above, discrimination between the 
classes is influenced both by the cluster-specific loading
matrices $\W_c$ and the noise precisions $\vec\tau_c$ that are allowed
to be different for each cluster. The former parameters provide for an easier
interpretation by directly inspecting the feature loadings.

\subsection{Inference}

For inference in our Bayesian model we adopt the Variational Bayesian approach \cite{attias1999inferring},
which is based on maximizing a lower bound on the log marginal likelihood of the data 
for a distribution that is of an easier form than the intractable true posterior distribution.
Typically, a factorized approximation $q(\Theta) = \prod_i q(\theta_i)$ is used,
where $\theta_i$ are some disjoint subsets of variables. It can be shown
that the optimal solution then is 
$q(\theta_i) \propto \exp(\E{\log p(\Theta, \X, \vec r)}_{\Theta \setminus \theta_i})$
in which the expectation is taken with respect to all variables except $\theta_i$.
For our model, we make the factorized approximation 
$p(\Theta | \X, \vec r) \approx q(\Theta) = q(\vec c)q(\vec\pi)q(\Z)q(\hat\W)q(\hat{\vec\alpha})\prod_{c=1}^Sq(\W_c^{(m)})q(\vec\tau_c)q(\gamma^c)q(\vec\alpha_c)$.

\section{Experiments with MEG data}
\begin{figure}[t]
\centering
\subfigure[ ]{%
\includegraphics[width=.42\textwidth]{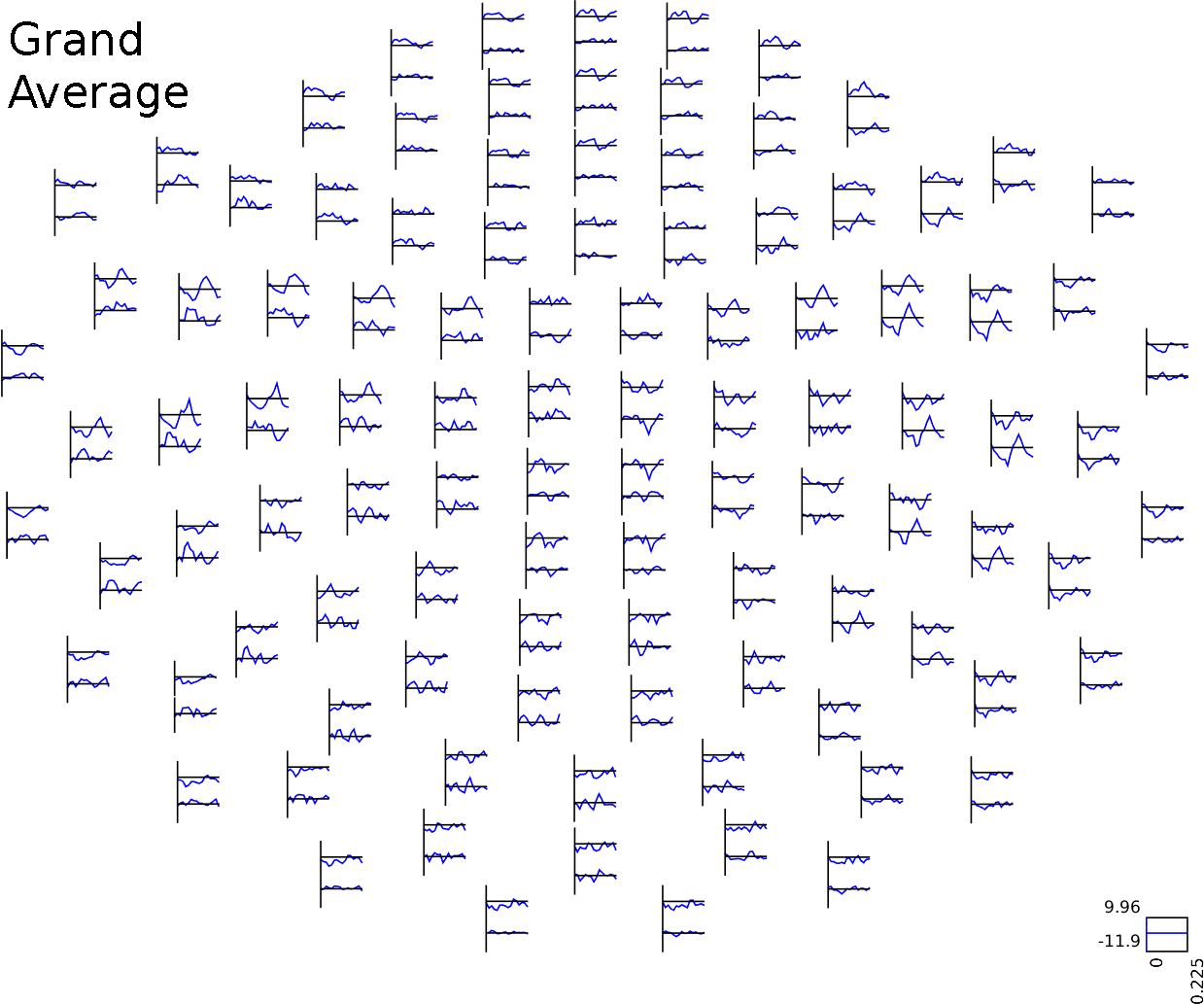}
\label{subfig:erp}}
\subfigure[ ]{%
\includegraphics[width=.35\textwidth]{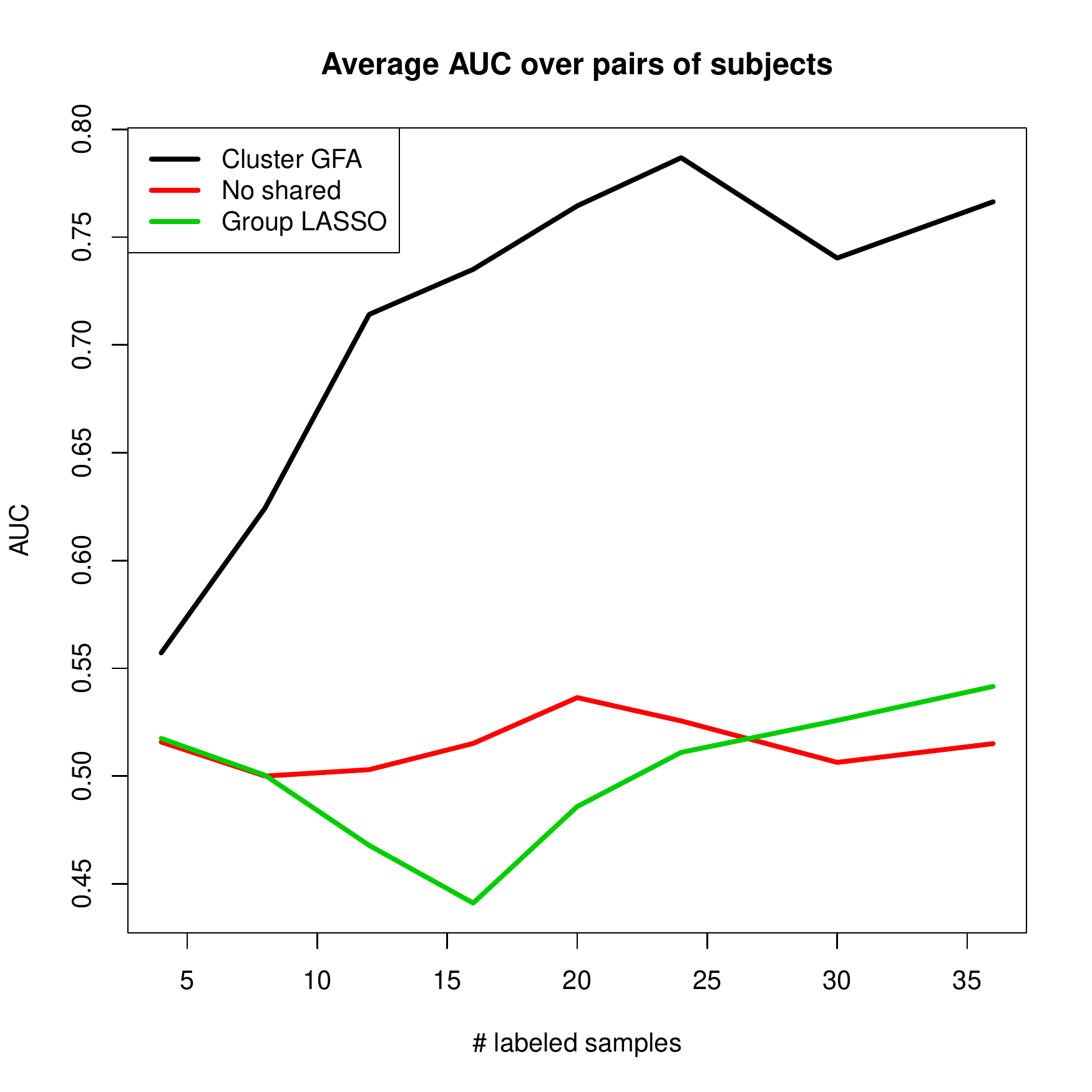}
\label{subfig:auc}}
\vspace{-.7em}
\caption{{\bf (a)} Average difference between the ``speak'' and ``listen'' conditions over the whole data (all subjects). Each pair of time series describes data of two MEG gradiometers at same location. Nose direction in the figure is upwards. {\bf (b)} Average AUC over the datasets plotted as a function of the size of the training data, for our model with and without the shared factors and  group LASSO. The LASSO begins slowly to gain performance only with the biggest sample sizes, while our model copes even with a very small data size.}
\label{fig:erp-auc}
\end{figure}

MEG data was collected simultaneously from two connected sites, each having a subject in an MEG device and a stimulus presentation computer, 
a system similar to those by \cite{baess2012meg} and \cite{zhdanov2015internet} using Elekta Neuromag 306 channel devices. 
The pairs of subjects engaged in a word game in which the two subjects took turns in uttering words to come up with a meaningful story. Data are available from 7 pairs of subjects. The lengths of the stories varied between 88--170 words. For our purposes we chose to use data from only one subject of each pair.
The data were preprocessed using SSS \cite{sss}, 
after which we discarded the magnetometer data leaving only the two gradiometers per sensor location, total of $M=204$ channels. 
The data were downsampled from 1000 Hz by a factor of 15 and high-pass filtered at 3 Hz to remove very slow signal changes and drifts.

Fig.~\ref{fig:erp-auc}\subref{subfig:erp}  depicts the conventional ERP analysis, comparing the listening and speaking conditions. We see that particularly in the auditory cortices (located on left and right sides slightly above the center) there exist
differences between the two conditions. Differences are also present at central or posterior
regions of the scalp.

\subsection{Results and interpretation}

We are classifying single trials, each trial consisting of either the subject listen a word or speak a word. 
We consider each MEG data channel as a data source or view. For a comparison method we use a Group LASSO implementation from \cite{breheny2015group} that sets hyperparameters by cross-validation.

As we have the whole single trial of one MEG gradiometer channel as one data source, the time series structure is modelled through the factor model.
More explicitly, the data matrix corresponding to source $m$ is
$\X^{(m)} = \begin{pmatrix}\x_1^{(m)}  \dots  \x_N^{(m)}\end{pmatrix}^T$, 
where $N$ is the number of trials, i.e. the number of spoken words.
Each $\x_n^{(m)}$ is a vector of time points within a 300 ms window.

\begin{figure}[t]
\centering
\includegraphics[width=.95\textwidth]{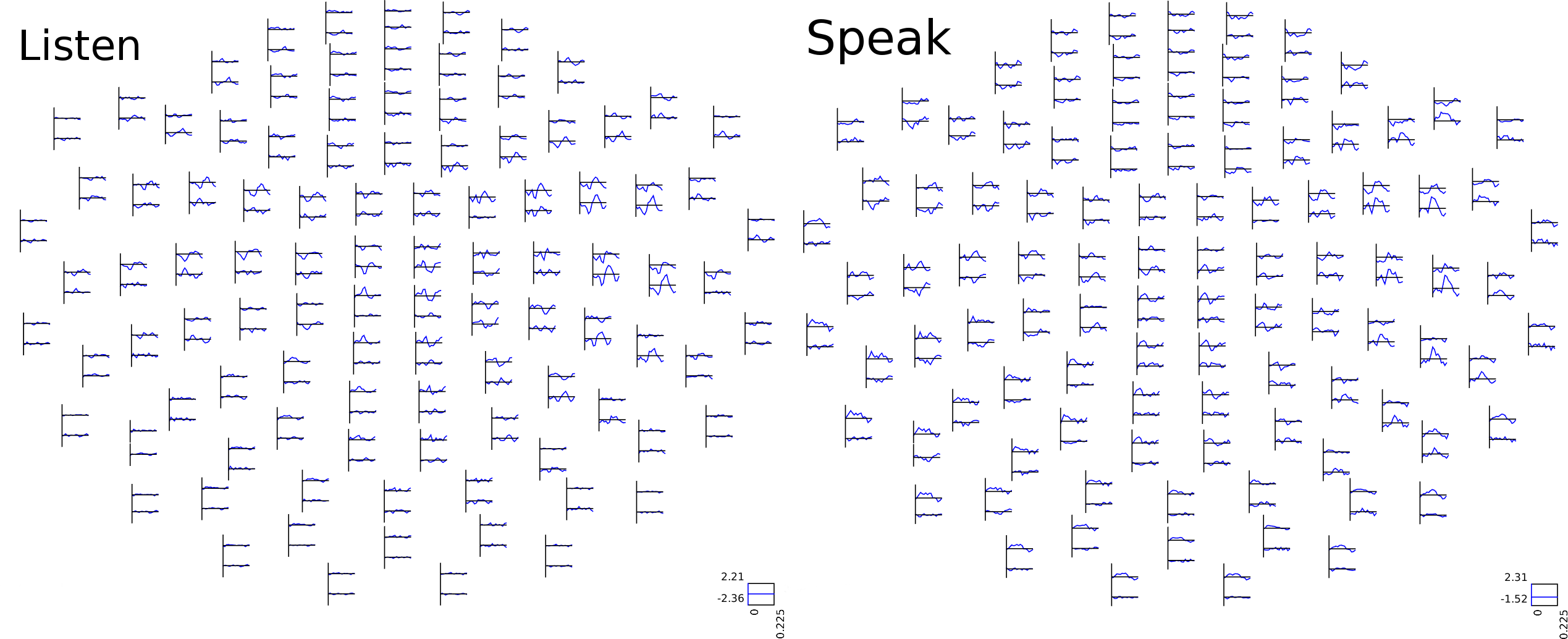}
\caption{Reconstructed data from factors in classes ``listen'' and ``speak,'' and the shared factors. The difference shows what parts of the total response are involved in explaining the class differences.}
\label{fig:my_label}
\end{figure}

We assess the performance of the classifiers with the AUC statistic and
report results based on a resampling technique, where training sets of a given size (see Fig.~\ref{subfig:auc}) and test sets of 10 samples were randomly drawn from the full datasets. We drew 10 pairs of training and test sets independently
for all subjects; pooling data from multiple subjects is not typically feasible with brain imaging data unless we can first align the data across subjects, see e.g. \cite{haxby13}. In the following results we vary the size of training data $N$ from 4 to 42. 

The results are presented in Fig.~\ref{fig:erp-auc}\subref{subfig:auc} for both
our model and group LASSO. Our model is clearly up to the task even with very small
training data, whereas group LASSO would require more data to improve performance.
Remarkably, a restricted version of our model without shared factors showed very low performance meaning that shared factors were significant.
For closer interpretation of the results, we computed our model also for the full datasets.
The grand average results in Fig.~\ref{fig:erp-auc}\subref{subfig:erp} give an idea about which areas of the brain are the most discriminative. As our model is generative, we can generate cluster-specific ERPs from the estimated model. The reconstruction of channel $m$ is calculated for cluster $c$ as $\hat\X^{(m)} = \Z\W_c^{(m)T}$
from which we average over the trials to obtain the ERPs shown in Fig.~\ref{fig:my_label}. We found that the reconstructions and the averages computed directly from data matched closely; the model found the existing differences and correctly picked them into the class-related clusters. 
With smaller training sets, the reconstructions were partial as fewer number of factors were in the model.

This simple case study demonstrated that the model is able to find discriminative signals even in single-trial MEG data. 
In this specific data the most discriminative signals are likely related to muscle activity which is present during speaking but not when listening; thus the model has picked up signs of this activity which is typically most visible in the channels closest to the edges of the MEG helmet. 
Also other areas are active in the ``speak'' condition. For the ``listen'' condition, both our model and the grand average show activations clearly in horizontally central areas that include the auditory cortex on both sides responsible for processing the word spoken by the other participant.

\section{Conclusion}

We introduced a classification model based on a mixture of group factor analyzers that share some of their
factors. The sharing seemed to be very significant in improving classification accuracy on our brain imaging
datasets; the model without shared factors performed much worse, as did the group LASSO baseline. 
In addition, we showed that our model gives interpretable results.  
The proposed model included two parameters that required tuning, namely the hyperprior parameter for the 
precisions of the shared loading matrices, and the weighting parameter for the Bernoulli distribution of the class labels.
We provided a simple way to set these, but it would be preferable to have a more rigorous analysis in future work.

{\scriptsize \subsubsection*{Acknowledgments. }
This work was financially supported by MindSEE (FP7 – ICT; Grant Agreement \#611570) and the Academy of Finland 
(CoE in Computational Inference Research COIN and LASTU).
The dual-MEG data set was collected in Brain2Brain project
funded by the European Research Council (Advanced Grant \#232946 to Riitta Hari,
Brain Research Unit, O.V. Lounasmaa Laboratory, Aalto University).
We thank P. Baess, R. Hari, T. Himberg, L. Hirvenkari, V. Jousmäki,
A. Mandel, J. Mäkelä, J. Nurminen, L. Parkkonen, and A. Zhdanov for
the possibility to use the anonymized dual-MEG data, A. Mandel for preprocessing of
the data set, and L. Hirvenkari for the stimulus timing.
Computational resources were provided by Aalto Science-IT project.\par}

\bibliographystyle{splncs03}
\bibliography{acml15}

\end{document}